\documentclass[pmlr]{jmlr}

\usepackage{longtable}
\usepackage{amssymb}

\usepackage{amsmath}
\usepackage{float}
\usepackage{graphicx} 
\usepackage{subcaption}

\usepackage{algorithm}
\usepackage{algorithmic}
\usepackage{graphicx}
\usepackage{subcaption}

\usepackage{booktabs}
\usepackage{algorithm}
\usepackage{algorithmic}
\usepackage{subcaption}
\usepackage{hyperref}
\usepackage{enumitem, xcolor}
\usepackage[load-configurations=version-1]{siunitx} 

\theorembodyfont{\upshape}
\theoremheaderfont{\scshape}
\theorempostheader{:}
\theoremsep{\newline}

\jmlrvolume{255}
\jmlryear{2024}
\jmlrworkshop{ML-DE Workshop at ECAI 2024}

\title[Time and State Dependent Neural Delay Differential Equations]{Time and State Dependent Neural Delay Differential Equations}

\author{\Name{Thibault Monsel} \Email{thibault.monsel@inria.fr}\\
    \addr LISN and INRIA, CNRS, Université Paris-Saclay, CNRS, 91405, Orsay, France
    \AND
  \Name{Onofrio Semeraro} \Email{onofrio.semeraro@universite-paris-saclay.fr} \\
  \Name{Lionel Mathelin} \Email{lionel.mathelin@cnrs.fr}\\
    \addr LISN, CNRS, Université Paris-Saclay, CNRS, 91405, Orsay, France
    \AND
   \Name{Guillaume Charpiat} \Email{guillaume.charpiat@inria.fr}\\
 \addr LISN, INRIA, Université Paris-Saclay, CNRS, 91405, Orsay, France}
 
\editors{Cecília Coelho, Bernd Zimmering, M. Fernanda P. Costa, Luís L. Ferrás, Oliver Niggemann}

\begin{document}

\maketitle

\begin{abstract}
Discontinuities and delayed terms are encountered in the governing equations of a large class of problems ranging from physics and engineering to medicine and economics. These systems cannot be properly modelled and simulated with standard Ordinary Differential Equations (ODE), or data-driven approximations such as Neural Ordinary Differential Equations (NODE). To circumvent this issue, latent variables are typically introduced to solve the dynamics of the system in a higher dimensional space and obtain the solution as a projection to the original space. However, this solution lacks physical interpretability. In contrast, Delay Differential Equations (DDEs), and their data-driven approximated counterparts, naturally appear as good candidates to characterize such systems. In this work we revisit the recently proposed Neural DDE by introducing Neural State-Dependent DDE (SDDDE), a general and flexible framework that can model multiple and state- and time-dependent delays. We show that our method is competitive and outperforms other continuous-class models on a wide variety of delayed dynamical systems. Code is available at the repository \href{https://github.com/thibmonsel/Time-and-State-Dependent-Neural-Delay-Differential-Equations}{here}.

\end{abstract}

\section{Introduction}
\label{sec:intro}

In many applications, one assumes the time-dependent system under consideration satisfies a Markov property; that is, future states of the system are entirely defined from the current state and are independent of the past. In this case, the system is satisfactorily described by an ordinary or a partial differential equation. However, the property of Markovianity is often only a first approximation to the true situation and a more realistic model would include past states of the system. Describing such systems has fueled the extensive development of the theory of delay differential equations (DDE) \citep{minorksy,Myshkis,Hale1963}. This development has given rise to many practical applications: in the modelling of molecular kinetics \citep{Roussel1996} as well as for diffusion processes \citep{Epstein1990DifferentialDE}, in physics for modeling semiconductor lasers \citep{laser}, in climate research for describing the El Nin\~o current \citep{nino,nino2}, infectious diseases \citep{cooper2020sir} and tsunami forecasting \citep{wu2022mining}, to list only a few.
\\

At the same time, the blooming of machine learning in recent years boosted the development of new algorithms aimed at modelling and predicting the behaviour of dynamical systems governing phenomena commonly found in a wide variety of fields. Among these novel strategies, the introduction of Neural Ordinary Differential Equations (NODEs) \citep{node} has contributed to further deepening the analysis of continuous dynamical systems modelling based on neural networks. NODEs are a family of neural networks that can be seen as the continuous extension of Residual Networks \citep{resnet}, where the dynamics of a vector $\mathbf{y}(t) \in \mathbb{R}^d$ at time $t$ -- hereafter often identified with the state of a physical system -- is given by the parameterized network $\mathbf{f}_{\theta}$ and the system's initial condition $\mathbf{y}_0$:
\begin{equation}
\frac{d \mathbf{y}(t)}{dt} = \mathbf{f}_{\theta}(t, \mathbf{y}(t)), \quad \mathbf{y}(0) = \mathbf{y}_0.
\end{equation}
NODEs have been successfully applied to various tasks, such as normalizing flows \citep{node_easy_to_solve,free_nf}, handling irregularly sampled time data \citep{latent_ode,kidger_irregular}, and image segmentation \citep{node_image_segmentation}.
\\

Starting from this groundbreaking work, numerous extensions of the NODE framework enabled to widen the range of applications. Among them, Augmented NODEs (ANODEs) \citep{dupont2019augmented} were able to alleviate NODEs' expressivity bottleneck by augmenting the dimension of the space allowing the model to learn more complex functions using simpler flows \citep{dupont2019augmented}. Let $\mathbf{a}(t) \in \mathbb{R}^p$ denotes a point in the augmented space, the ODE problem is formulated as 
\begin{equation}
\begin{aligned}
\frac{d}{dt}
\begin{bmatrix} 
	\mathbf{y}(t)\\
	\mathbf{a}(t)
\end{bmatrix} = \mathbf{f}_{\theta} 
\left(t,
\begin{bmatrix} 
	\mathbf{y}(t)\\
	\mathbf{a}(t)
\end{bmatrix}
 \right), \quad 
\begin{bmatrix} 
	\mathbf{y}(0)\\
	\mathbf{a}(0)\\
\end{bmatrix} = 
\begin{bmatrix} 
	\mathbf{y}_0\\
	\mathbf{0}
\end{bmatrix}.
\end{aligned}
\end{equation} 
%
By introducing this new variable $\mathbf{a}(t)$, ANODE overcomes the inability of NODE to represent particular classes of systems. However, this comes with the cost of augmenting the data into a higher dimensional space, hence losing interpretability. Among the alternative techniques proposed for circumventing the limitations rising from the modelling of non-Markovian systems, the Neural Laplace model \citep{neural_laplace} proposes a unified framework that solves differential equations (DE): it learns DE solutions in the Laplace domain. The Neural Laplace model cascades 3 steps: first, a network $\mathbf{h}_{\gamma}$ encodes the trajectory, then the so-called Laplace representation network $\mathbf{g}_{\beta}$ learns the dynamics in the Laplace domain to finally map it back to the temporal domain with an inverse Laplace transform (ILT). With the state $\mathbf{y}$ sampled $T$ times at arbitrary time instants, $\mathbf{h}_{\gamma}$ gives a latent initial condition representation vector $\mathbf{p} \in \mathbb{R}^K$:
\begin{equation}
\begin{aligned}
& \mathbf{p} = \mathbf{h}_{\gamma}((\mathbf{y}(t_1), t_1), \dots, (\mathbf{y}(t_T), t_T)),
\end{aligned}
\end{equation}
that is fed to the network $\mathbf{g}_{\beta}$ to get the Laplace transform 
\begin{equation}
\begin{aligned}
& \mathbf{F}(\mathbf{s}) = v\left( \mathbf{g}_{\beta}(\mathbf{p}, u(\mathbf{s})) \right),
\end{aligned}
\end{equation}
with $u$ a stereographic projector and $v$ its inverse. Ultimately, an ILT step is applied to reconstruct state estimate $\widehat{\mathbf{y}}$ from the learnt $\mathbf{F}(\mathbf{s})$. 
\\

As an alternative to the aforementioned techniques, one may directly address the DDE problem by working within the framework of neural network-based DDEs. Despite the success of the NODEs philosophy, the extension to DDEs has barely been studied yet, possibly owing to the challenges of using general purpose DDE solvers. DDEs extend ODEs by incorporating additional terms into their vector fields, which are states delayed by a certain time $\tau$. Recently, \cite{first_neural_dde} introduced a neural network based DDE with one single constant delay:
\begin{equation}
\begin{aligned}
& \frac{d \mathbf{y}(t)}{dt} = \mathbf{f}_{\theta}(t, \mathbf{y}(t), \mathbf{y}(t-\tau)), \quad \tau \in \mathbb{R}^{+} \\
& \mathbf{y}(t<0) = \boldsymbol{\phi}(t),
\end{aligned}
\label{eqn:first_neural_dde}
\end{equation}
where $\boldsymbol{\phi}(t)$ is the system's history function, $\tau$ a constant delay and $ \mathbf{f}_{\theta}$ a parameterized network. This work was next extended in the Neural Piece-Wise Constant Delays Differential Equations (NPCDDEs) model in \cite{neural_piece_wise_dde}. Compared to NODE and its augmented counterpart, neural network-based DDEs do not require an augmentation to a higher dimensional space in order to be a universal approximator, thus preserving physical interpretability of the state vector and allowing the identification of the time delays. Nonetheless, the current variants of Neural DDE models only deals with a single constant delay or several piece-wise constant delays, thus lacking the generalization to arbitrary delays. Moreover, to the best of our knowledge, no machine learning library or open-sourced code exists to model not only these very specific types of DDEs but also any generic DDEs. 
\\

In this work, we introduce Neural State-Dependent DDE (SDDDE) model: an open-source, robust python DDE solver compatible with neural networks. Neural SDDDE is based on a general framework that pushes the envelope of Neural DDEs by handling DDEs with several delays in a more generic way. The implementation further encompasses general time- and state-dependent delay systems which extends the reach of \cite{neural_piece_wise_dde}. 
\\

In the remainder of the paper, we briefly introduce the framework in Sec.~\ref{sec:dde}. Implementation and methodologies are further discussed in Sec.~\ref{sec:methods}. Experiments and comparisons with the state-of-the-art techniques are detailed in Sec.~\ref{sec:experiments}, using as benchmark numerous time-delayed models of incremental complexity. Our model is shown below to compare favorably with the current models on DDEs systems. Conclusions and outlook finalize the article in Sec.~\ref{sec:conclusions}.

\section{Neural State-Dependant DDEs}\label{sec:dde}

In this section, we introduce the Neural State-Dependent Delay Differential Equation (SDDDE) model, which is designed to accommodate various types of delays, including constant, time-dependent, and state-dependent delays. However, it is important to emphasize that this approach cannot handle delays that are continuous, meaning those expressed through integrals, as typically found in integro-differential equations. The specific types of delays that our model can accommodate are summarized in Table \ref{table_delays}. This table provides a clear overview of the capabilities of the SDDDE model regarding different delay types. A generic Delay Differential Equation is described by:
\begin{equation}
\begin{aligned}
 & \frac{d \mathbf{y}(t)}{dt} = \mathbf{f}_{\theta}(t, \mathbf{y}(t), \mathbf{y}(t-\tau_1(t, \mathbf{y}(t))), \dots, \mathbf{y}(t-\tau_k(t, \mathbf{y}(t)))) \\
&  \mathbf{y}(t<0) = \mathbf{\phi}(t),
\end{aligned}
\label{eqn:dde}
\end{equation}
where $\mathbf{\phi}: \mathbb{R}^- \rightarrow \mathbb{R}^d$ is the history function, $\tau_i: \mathbb{R} \times \mathbb{R}^d \rightarrow \mathbb{R}^{+}$  a delay function and $\mathbf{f}_{\theta}:[0, T] \times \mathbb{R}^d \times \dots \times \mathbb{R}^d \rightarrow \mathbb{R}^d$ a parameterized network. In Appendix \ref{ap:dde_appendix}, we provide more general and detailed information on DDEs, how we integrate them and discuss memory and time complexities for Neural SDDDE.

\begin{table}
\caption{Existing works that deal with DDEs. Our implementation deals with a wider range of delays ($g$ is a scalar-valued function)}
\label{table_delays}
\vskip 0.15in
\begin{center}
\begin{small}
 \resizebox{16cm}{!}
{
\begin{tabular}{lcccccc}
\toprule
Delay types & $\tau$'s Definition & Neural DDE & NPCDDEs & Neural Laplace & Neural SDDDE \\
\midrule
References & - & \citet{first_neural_dde} & \citet{neural_piece_wise_dde} & \citet{neural_laplace} & This work \\
Constant & $\tau = a$ & $\surd$ & $\times$ & $\surd$ & $\surd$ \\
Piece-wise constant & $\tau = \left\lfloor \frac{t-a}{a} \right\rfloor a$ & $\times$ & $\surd$ & $\surd$ & $\surd$ \\
Time-dependent & $\tau = g(t)$ & $\times$ & $\times$ & $\surd$ & $\surd$ \\
State-dependent & $\tau = g(t, y(t))$ & $\times$ & $\times$ & $\times$ & $\surd$ \\
Continuous & $\tau = \int_0^{t}g(s, y(s)) \mathrm{d}s$ & $\times$ & $\times$ & $\times$ & $\times$ \\
\bottomrule
\end{tabular}
}
\end{small}
\end{center}
\end{table}

\section{Methods}\label{sec:methods}

In the following, we discuss similarities, drawbacks and benefits of Neural SDDDE compared with the following models: NODE, ANODE and Neural Laplace. We recall that Neural SDDDE is a direct method for solving DDEs, specifically designed to handle delayed systems. This is not the case for ANODE where flexibility is obtained by introducing a higher dimensional space. Lastly, Neural Laplace can solve a broader class of differential equations, although with some limitations that we pinpoint in the following.
\\

From the theoretical viewpoint, the Laplace transformation is often a tool used in proofs on DDEs \citep{Cook_book_ref_dde} since it allows to transform linear functional equations in $\mathbf{f}(\mathbf{y}(t))$ involving derivative and differences into linear equations involving only $\mathbf{F}(\mathbf{s})$. Thus, time-dependent and constant delay DDEs are transformed into linear equations of $\mathbf{F}(\mathbf{s})$ using the Laplace transform. This transformation enables Neural Laplace to bypass the explicit definition of delays, whereas Neural SDDDE needs the delays to be specified unless the vector flow $\mathbf{f}$ and delays are learnt jointly. These observations tie Neural Laplace to Neural SDDDE as they can be seen as similar models but living in different domains. However, a limitation of Laplace transformation-based approaches is that they are not defined for DDEs with state-dependent delays, thus restricting the class of time-delayed equations that one can solve with this technique. 
\\

Neural Laplace is a model that needs memory initialized latent variables, i.e., a long portion of the solution trajectory needs to be fed in order to get a reasonable representation of the latent variable. In contrast, by design, NODE and ANODE require information only at the initial time to predict the system dynamics. From this viewpoint, Neural SDDDE lies between these two frameworks as it requires a history function $\boldsymbol{\phi}(t)$ to be provided for $t\in[-\tau_{\mathrm{max}},0]$, where $\tau_{\mathrm{max}}$ is the maximum delay encountered during integration. This is more demanding than solely relying on an initial condition $\mathbf{y}(0)$ but far less than memory-based latent variables methods. Moreover, the training and testing schemes for Neural Laplace is constrained by this very same observation since future events can only be predicted after a certain observation time. This also makes the length of the trajectory to feed to Neural Laplace a hyperparameter to tune. This is not the case with NODE, ANODE and our approach. 
\section{Experiments}\label{sec:experiments}

We evaluate and compare the Neural SDDDE on several dynamical systems (time, state-dependent and constant delays) listed below coming from biology and population dynamics. We show that Neural SDDDE outperforms a variety of continuous-depth models and demonstrates its capabilities in simulating delayed systems. For all the systems listed in this section, data generation information is gathered in Appendix~\ref{ap:data_generation}.



\subsection{Description of the test cases}

\paragraph{Time-dependent delay system} Here, we study a time-dependent delayed logistic equation \citep{arino2006alternative}:
\begin{equation}
\begin{aligned}
&\frac{d y(t)}{dt} = y(t) \bigl[ 1 -y(t-\tau(t))\bigr],
\end{aligned}
\end{equation}
with $\tau(t) = 2 + \sin(t)$. We integrate in the time range $[0,20]$ and define the constant history function $\phi(t) = x_0$, where $x_0$ is sampled uniformly from $[0.1,2.0]$.

\paragraph{State-dependent time-delay system} In this example, we consider the 1-D state-dependent Mackey Glass system from \citet{state_dependent_mg} with a state-dependent delay:
\begin{equation}
\frac{d y(t)}{dt} = -\alpha(t) y(t) + \beta(t) \frac{ y^2(t - \tau(y))}{1 + y^2(t - \tau(y))} + \gamma(t)
\end{equation}
with $\alpha(t) = 4 + \sin(t) + \sin(\sqrt{2t}) + \frac{1}{1 + t^2}$, $\beta(t) = \gamma(t) = \sin(t) + \sin(\sqrt{2t}) + \frac{1}{1 + t^2}$ and $\tau(y) = \frac{1}{2} \cos(y(t))$. The model is defined on the time range $[0,10]$ and the constant history function is $\phi(t) = x_0$ with $x_0$ sampled uniformly from $[0.1,1]$.

\paragraph{Delayed Diffusion Equation} Finally, we choose the delayed PDE taken from \citet{arino2009delay}. Such dynamics can for example model single species growth in a food-limited environment.
\begin{equation}
\begin{aligned}
&\frac{\partial u}{\partial t}(x,t) = D \frac{\partial^2 u}{\partial x^2}(x,t) + r u(x,t) \left( 1 - u(x,t-\tau) \right),
\end{aligned}
\end{equation}
where $D=0.01, r=0.9$ and $\tau=2$. We integrate in the time range $[0,4]$, the spatial domain is $\mathcal{D}_x = [0,1]$ with periodic boundary conditions and define the history function $\phi(x, t) = a \sin(x) e^{-0.01t} $ where $a$ is uniformly sampled from $[0.1,4.0]$. The spatial domain is discretized with a uniform grid of resolution $\Delta x=0.01$.

\subsection{Evaluation}
\label{ss:evaluation}
We assess the performance of the models with their ability to predict future states of a given system. The metric used is the mean square error (MSE) in all cases. Neural Laplace predicts only after a burn-in time since a part of the observed trajectory is used to learn a latent initial condition vector $\mathbf{p}$. Since NODE, ANODE and Neural SDDDE can be seen as initial value problems (IVPs) we produce trajectories from initial conditions and compute the MSE with respect to the whole trajectory.
On each DDE system, to assess the quality of each model, we elaborate additional experiments alongside with the \textit{test set predictions} that can be found in Appendix \ref{ap:step_history_experiments}.
\\

As a reminder, to produce outputs, NODE and ANODE need an initial condition, Neural SDDDE the history function and Neural Laplace a portion of the trajectory. To ensure a fair comparison in our experiments, we opt to give Neural Laplace the same information as Neural SDDDE, specifically the history function.
For one of the models presented above, the Neural Laplace method is given a much larger chunk of the trajectory, in accordance with what its authors were considering.

\subsection{Results}

Test errors for each dynamical system are reported in Table~\ref{test_rmse}. Complementary information are included in the Appendix \ref{ap:training_curves} for what concerns the training process; model and training hyperparameters.


\begin{table}[H]
\begin{center}
\begin{small}
\resizebox{14cm}{!}{
\begin{tabular}{llll}
\toprule
& Time Dependent DDE & State-Dependent DDE & Delay Diffusion \\
\midrule
NODE                & $.72 \pm .086$      & $.0355 \pm .00064$       & $.0029 \pm .0014$ \\
ANODE               & $.00962 \pm .00368$ & $.00011 \pm .000071$     & $.00087 \pm .00035$ \\
Neural Laplace      & $.00191 \pm .0006$  & $.00049 \pm .00078$      & $\mathbf{.00064 \pm .00016}$ \\
Neural SDDDE        & $\mathbf{.000989 \pm  .00017}$ & $\mathbf{.0000215 \pm .00001}$ & $.00075 \pm .00019$ \\
\bottomrule
\end{tabular}
}
\end{small}
\caption{Test MSE averaged over 5 runs (random model initialization seed) of each experiment with their standard deviation. Best result bolded.}
\label{test_rmse}
\end{center}
\end{table}



\begin{figure}[h]
    \centering
    \begin{minipage}{0.45\textwidth}
        \centering
        \includegraphics[width=\textwidth]{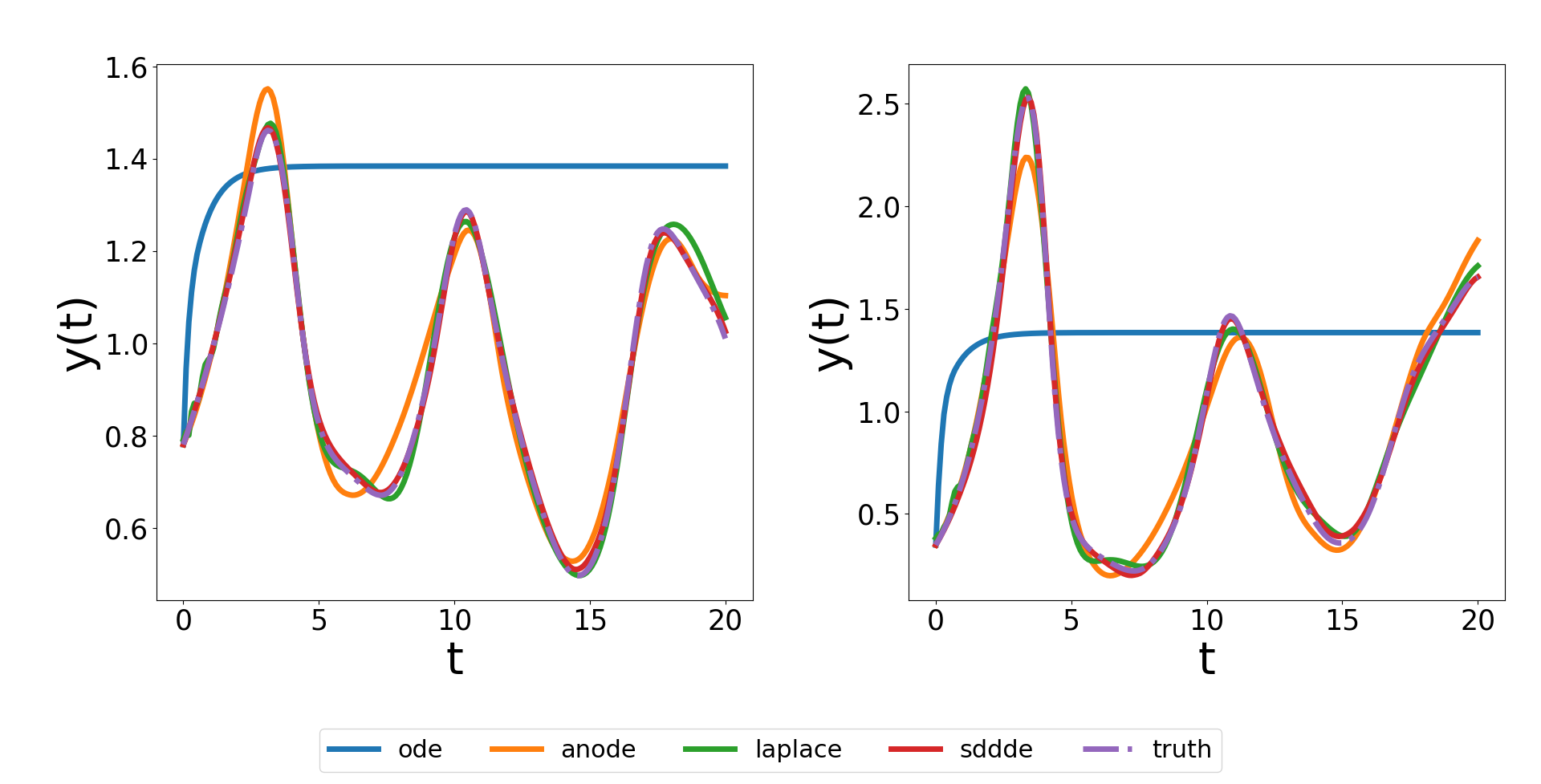}
         \caption{Time Dependent DDE randomly sampled test trajectory plots }\label{subfig:time_testset}
    \end{minipage}\hfill
    \begin{minipage}{0.45\textwidth}
        \centering
         \includegraphics[width=\textwidth]{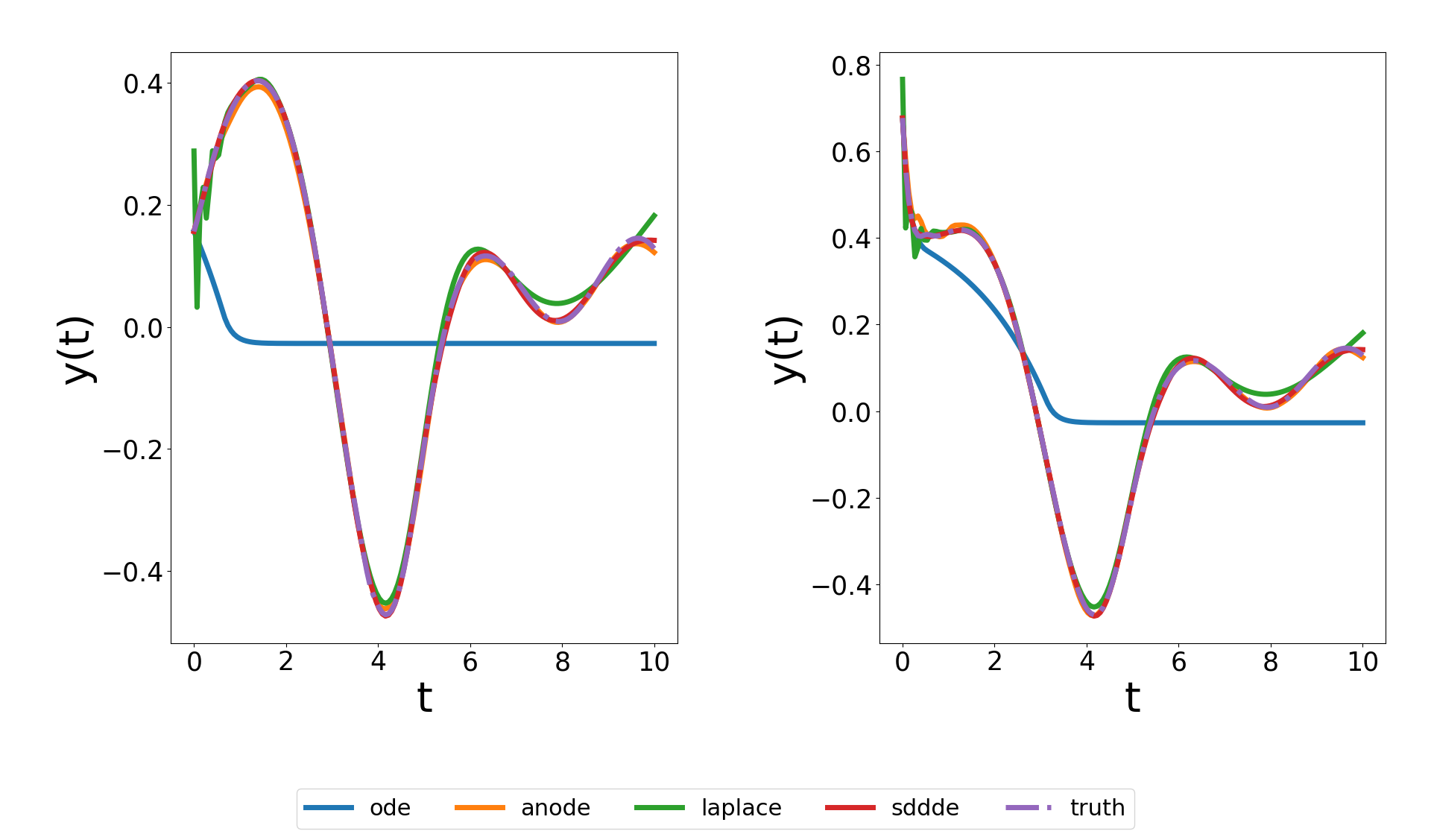}
         \caption{State Dependent DDE randomly sampled test trajectory plots }\label{subfig:state_testset}
    \end{minipage}
\end{figure}

 \begin{figure}[h]
    \centering
    \includegraphics[width=\textwidth, trim={0cm 2.5cm 0 8cm},clip]{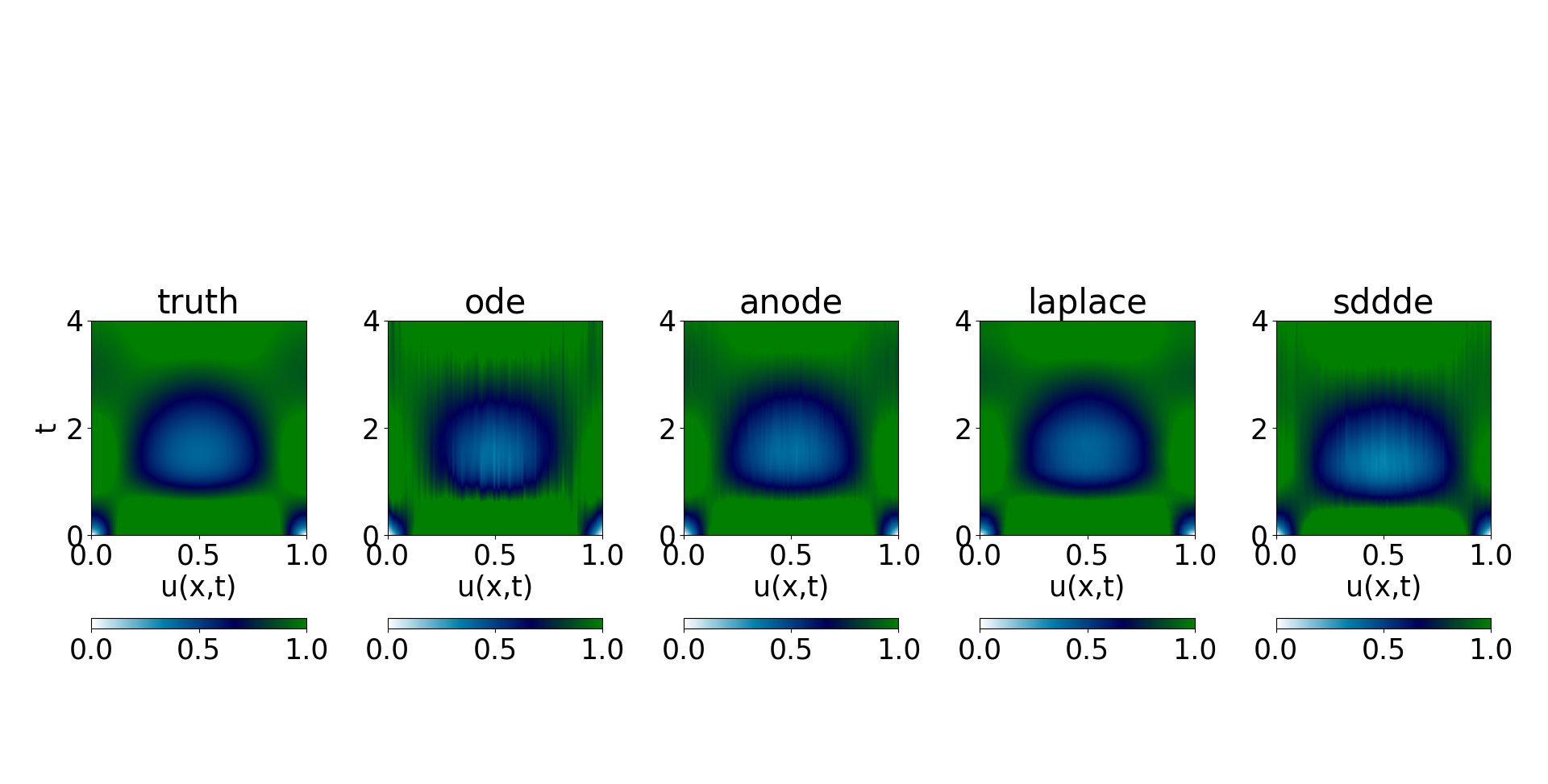}\caption{Diffusion Delay PDE  randomly sampled from the testset}\label{subfig:diffusion_testset}

    \vspace{1cm}

    \includegraphics[width=\textwidth, trim={0cm 2.5cm 0 8cm},clip]{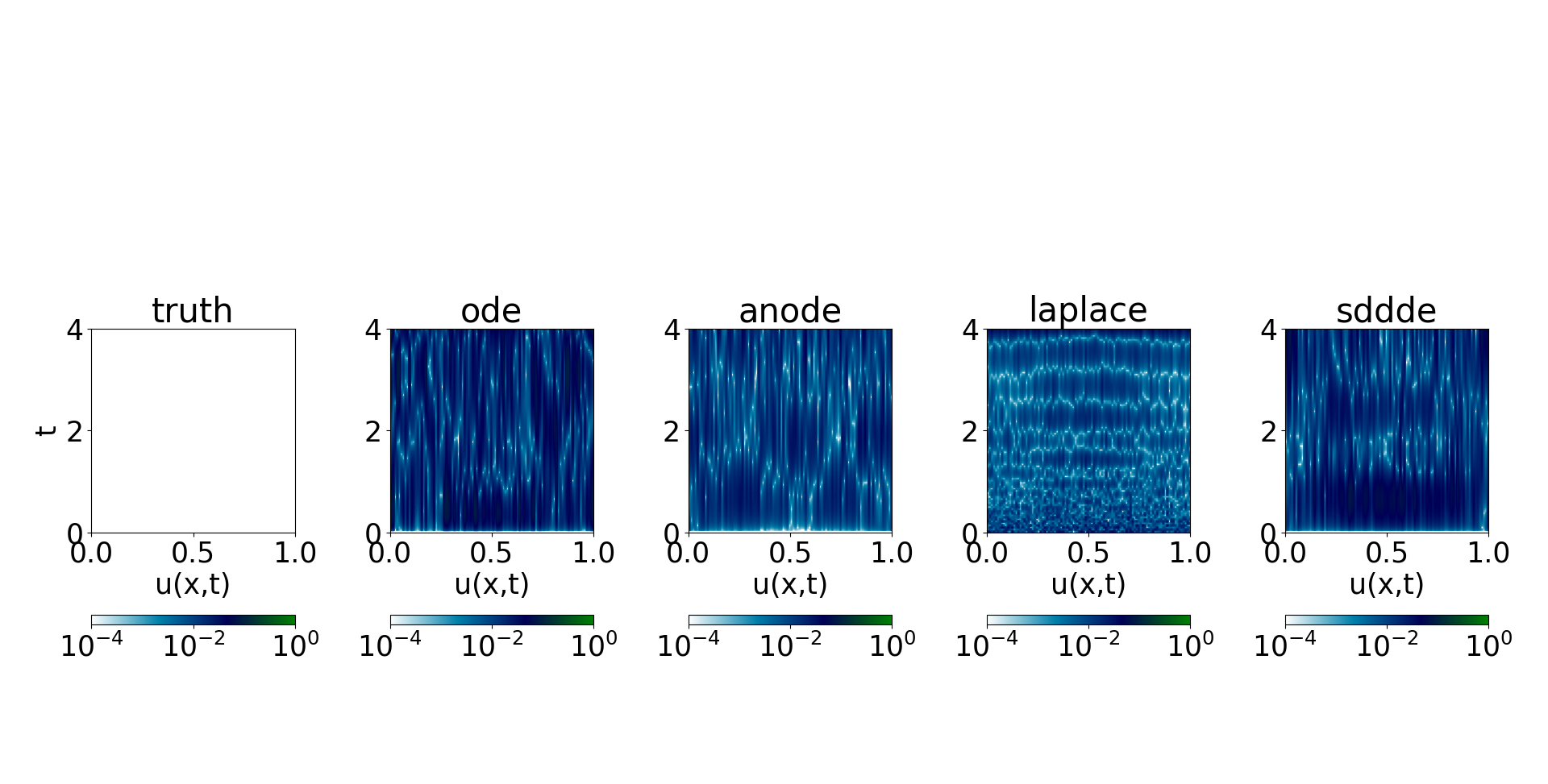}
    \caption{Absolute error of Diffusion Delay PDE randomly sampled from the testset}\label{subfig:error_diffusion_diff_testset}
    
\end{figure}

\paragraph{Testset prediction} Neural SDDDE almost consistently outperforms all other models across the DDE systems discussed in Section \ref{sec:experiments}, as demonstrated in Figures \ref{subfig:time_testset}, \ref{subfig:state_testset}, \ref{subfig:diffusion_testset}, and \ref{subfig:error_diffusion_diff_testset}. Neural Laplace appears to suffer from the Runge phenomenon (such a phenomenon is a
problem of oscillation at the edges of an interval that occurs when using polynomial interpolation with polynomials of high degree over a set of equispaced interpolation
points), particularly evident in the State-Dependent DDE (Figure \ref{subfig:state_testset}). This issue likely stems from the ILT algorithm providing too few query points. As expected, Neural ODE is the most limited model, generally predicting only the mean trajectory of the dynamical systems being considered. ANODE yields satisfactory results, with the exception of the Time Dependent DDE (Figure \ref{subfig:time_testset}). For the Diffusion Delay PDE, all models predict the PDE's evolution with an absolute error reaching up to $10^{-2}$, as shown in Figure \ref{subfig:diffusion_testset}. The absolute error, depicted in Figure \ref{subfig:error_diffusion_diff_testset}, illustrates the discrepancies between the models. Neural Laplace produces more errors across the entire spatial domain for given time steps, while IVP models have errors localized in specific spatial regions.

\begin{figure}[H]
\centering

\begin{minipage}{0.5\textwidth}
    \centering
    \includegraphics[width=\textwidth,height=\textheight, keepaspectratio]{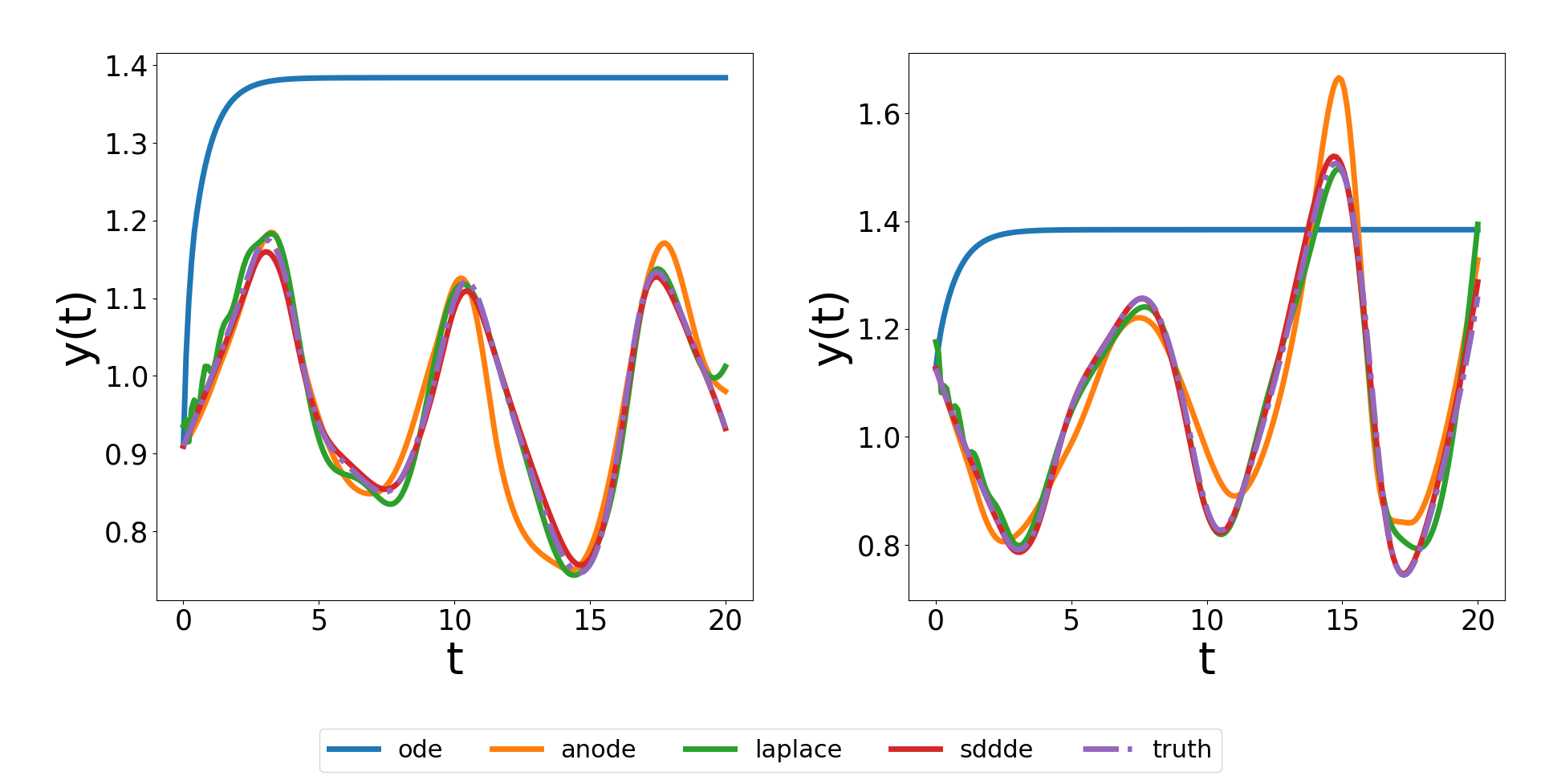}
    \caption{Time-dependent DDE randomly sampled testset trajectories where 50\% of data is fed to Neural Laplace}
\label{fig:50_neural_laplace_comparison}
\end{minipage}
\hfill 
\begin{minipage}{0.45\textwidth}
    \centering
\begin{table}[H]
\begin{center}
\begin{small}
\begin{tabular}{lc}
\toprule
& Test MSE \\
\hline
Neural Laplace & $.00125 \pm .000798$ \\
Neural SDDDE & $\mathbf{.000989 \pm  .00017}$ \\
\bottomrule
\end{tabular}
\end{small}
\caption{Time Dependent test MSE averaged over 5 runs with their standard deviation. Best result bolded.}
\label{50_laplace_mse}
\end{center}
\end{table}

\end{minipage}

\end{figure}

\paragraph{Increasing trajectory fed for Neural Laplace} Instead of providing the same history as for Neural SDDDE, Neural Laplace is now provided 50\% of the trajectory to build its latent initial condition representation vector $\mathbf{p}$. To that purpose, the first half of the trajectory is used in Neural Laplace to predict the second half. We choose to train the model on the Time Dependent DDE along with the same training procedure (see Appendix~\ref{ap:training_curves}). Provided the test MSE of Table \ref{test_rmse}, we choose to only compare the test MSE of Neural SDDDE and Neural Laplace in Table \ref{50_laplace_mse}. By comparing Figure \ref{fig:50_neural_laplace_comparison} and \ref{subfig:time_testset} one can clearly note that the Runge phenomenon is almost absent and predictions are almost as good as Neural SDDDE. This confirms that, in general, Neural Laplace needs more than the history function in order to correctly simulate DDEs. 

\section{Conclusion and Future Work}\label{sec:conclusions}

In this paper, we introduced Neural State-Dependent Delays Differential Equations (Neural SDDDE) capable of solving DDEs with any type of delays via neural networks. This open-source, robust python DDE solver compatible with neural networks pushes the current envelope of Neural DDEs by handling the delays in a more generic way. To the best of our knowledge, no machine learning library or open-sourced code is available to model such a large class of DDEs. 

To validate the effectiveness of Neural SDDDE, we conducted a series of benchmark tests, comparing it against NODEs, the augmented version ANODE, and Neural Laplace. These numerical experiments covered a diverse range of models, including time- and state-dependent scenarios, as well as a delayed Partial Differential Equation (PDE). Our findings revealed that Neural SDDDE accurately reproduced the dynamics across all scenarios tested. Furthermore, Neural SDDDE demonstrated superior performance in terms of accuracy and reliability when compared to the other established methods.

We believe this flexible and versatile tool may provide a valuable contribution to several fields such as control theory where time-delay are often considered. In particular, it may prove useful in learning a model for partially observed systems whose dynamics of observables can be learned, under mild conditions, from their time-history.

\acks{This research is funded by the grant from Agence Nationale de Recherche : Project number ANR-20-CE23-0025-01.}

\bibliography{references}

\appendix

\appendix\label{sec:appendix}
\newpage


\section{Overview in DDE integration}\label{ap:dde_appendix}

This Appendix section is a self-contained introduction on DDE integration.

\subsection{Definition}

We recall the definition: a delay differential equation (DDE) is defined by
\begin{equation}
\begin{aligned}
 & \frac{d \mathbf{y}(t)}{dt} = \mathbf{f}_{\theta}(t, \mathbf{y}(t), \mathbf{y}(t-\tau_1), \dots, \mathbf{y}(t-\tau_k)) \\
& \tau_i = \tau_i(t, \mathbf{y}(t)), \quad \forall i \in \left\{1, 2, \ldots,  k\right\} \\
&  \mathbf{y}(t<0) = \boldsymbol{\phi}(t),
\end{aligned}
\label{eqn:dde_def}
\end{equation}
where $\mathbf{\boldsymbol{\phi}}: \mathbb{R}^- \rightarrow \mathbb{R}^d$ is the history function, $\tau_i: \mathbb{R} \times \mathbb{R}^d \rightarrow \mathbb{R}$ a delay function and $\mathbf{f}_{\theta}:[0, T]^{k} \times \mathbb{R}^d \rightarrow \mathbb{R}^d$ can be a parameterized network. 
\\

Some problems can arise in DDEs that can cause numerical difficulties. First, breaking points may occur in various derivatives of the solution $\mathbf{y}$. Second, a delay may vanish, i.e., $\tau_i \rightarrow 0$. The first difficulty is due to the presence of delays terms. In general, DDEs possess a derivative jump (or discontinuity, breaking point) at the initial time point $t=0$ because 

\begin{equation*}
    \phi^{\prime}(t=0^{-}) \neq  \mathbf{y}^{\prime}(t=0^{+}) 
\end{equation*}

Moreover, the history function $\phi$ may also have discontinuities too. Discontinuities can then arise and propagate from the history function and initial point in $\mathbf{y}$ or its higher derivatives \citep{numerical_dde}. The second issue may force the solver to take too many small steps. \cite{numerical_dde} transforms the DDE problem into a discontinuous initial value problem (IVP). Alike ODEs, existence and uniqueness theorems for DDEs are based on the continuity of the functions with respect to $t$ and Lispchitz continuity with respect to $\mathbf{y}$ and its delayed counterparts $\mathbf{y}(t-\tau(t,\mathbf{y}))$. For constant, time-dependent and state-dependent delays, these problems have been widely investigated by \citet{Cook_book_ref_dde, existence_general_delay, Driver1962} and \citet{hale2006functional}.

\subsection{Example sketch of integrating a simple DDE}

Let us consider a first order DDE with a constant time delay $\tau$ and a constant history function $\phi(t) = \mathbf{y}_0$. In the most general case, we have our first discontinuity at $t=0$ since $ \phi^{\prime}(t=0^{-}) \neq  \mathbf{y}^{\prime}(t=0^{+})$, so we need to be careful on integration on these derivative jumps.

\begin{equation}
    \frac{d \mathbf{y}(t)}{dt}  = f(t, \mathbf{y}(t), \mathbf{y}(t-\tau)), \quad \text{with} \quad \mathbf{y}(t<0) =  \mathbf{y}_0
\label{eqn:simple_dee}
\end{equation}

On the time interval $t \in ]0;\tau[$ there are no discontinuities and the DDE becomes 

\begin{equation*}
    \frac{d \mathbf{y}(t)}{dt}  = f(t, \mathbf{y}(t), \mathbf{y}_0), \quad \text{with} \quad \mathbf{y}(0) = \mathbf{y}_0
\end{equation*}

This formulation problem reshapes the DDE problem into an ODE one that we know how to solve with ease. Let us introduce the interpolated function $\phi_1(t)$ to be the solution of the DDE of Equation \ref{eqn:simple_dee} on the interval $]0;\tau[$.  \\

On the time interval $t \in ]\tau;2\tau[$ there are no discontinuities and the DDE becomes 

\begin{equation*}
     \frac{d \mathbf{y}(t)}{dt}= f(t,  \mathbf{y}(t), \phi_1(t)), \quad \text{with} \quad \mathbf{y}(\tau) = \phi_1(\tau) 
\end{equation*}

Once again we have an ODE on this interval. Iteratively, we can solve the DDE on successive intervals and this will yield a piecewise continuous solution because of the initial point discontinuity. \\

One might have noticed that during an integration step, we need the interpolated function of $ \mathbf{y}(t-\tau)$. This means that the DDE method is based on the \textit{continuous extensions} of numerical ODE schemes.

\subsection{Discontinuity tracking}

In the general case, discontinuity that arise from the delay terms are not known a priori unless we are dealing with constant delays \citep{constant_delay_dde}. During each integration step of our DDE, one must check for discontinuities by checking the roots of the following functions $g_{is}$ \citep{numerical_dde}. Let us stipulate that we integrate from $t_n$ to $t_{n+1}$ and the previous detected discontinuities are $\{ \lambda_{-m}, \dots, \lambda_0, \dots, \lambda_{r-1} \}$ where the first $m+1$ jumps $\{ \lambda_{-m}, \dots, \lambda_0 \}$ are given by the history function and the initial point and the rest were found during previous integration steps. Let 

\begin{equation}
    \forall (i,s) \in [0, \dots, r] \times [-m, \dots,r-1],\quad g_{is}(t) = t - \tau_i - \lambda_s
\end{equation}

The new discontinuity $\lambda_r$ is defined as

\begin{equation}
    \lambda_r = min \{\lambda > \lambda_{r-1},  \lambda \text{ is a root of odd multiplicity of } g_{is}(t) \}
\end{equation}

If $\lambda_r$ is null then the integration step is valid otherwise you redo one from $t_n$ to $\lambda_r$. A detailed algorithm procedure is given in \cite{numerical_dde}. \\

The first to do such an iterative process to find the discontinuities $\lambda_r$ and modify the integration step bounds is \cite{discontinuity_tracking}. An alternative approach relies on stepsize control was proposed by \cite{first_time_discont_tracking} and \cite{neves}. These methods, give up on tracking the discontinuities, which are instead assumed to be automatically included by estimating the error of the integration step. A rejected step will result in a detection of a discontinuity jump and this is the default implementation done in Julia  \textit{DelayDiffEq} package \citep{julia_dde}. \\

For an ODE method of order $p$, we usually ask the solution to be at least $\mathcal{C}^{p+1}$ continuous. Therefore, to have a successful integration, it is crucial to include in the mesh of points all of the discontinuity of $y^{(k)}$ at least for $k \leq p+1$. Consequently, discontinuity tracking needs to abide by these rules.

\subsection{Unconstrained time stepping}

Regardless of the method chosen to integrate a DDE; relying on the error estimate of the stepsize method or tracking the breaking points, being able to take arbitrarily large steps that are suggested by the numerical solver is a nice to have. This means that sometimes the formulation of our problem becomes implicit because our approximation solution $\mathbf{y}$ applied to all delays terms in our integration step is simply not yet known. This makes the overall method implicit even if the discrete method we are using is explicit. We call this occurence overlapping, \cite{numerical_dde} shows that the issue at hand is well defined and solvable for time and state dependent delays. Let us briefly describe the algorithmic procedure when we are dealing with overlapping (ie $t_{n+1}-t_n >\tau$). Given equation \ref{eqn:simple_dee} and an integration step from $t_n$ to $t_{n+1}$. $\mathbf{y}(t-\tau)$ is at the very best partially known and need to be extrapolated to get an good approximation of $\mathbf{y}(t_{n+1})$. The following actions are taken : 
\begin{itemize}[noitemsep,topsep=0pt] 
    \item Choose an initial guess for the interpolant $\Pi_n$ of $\mathbf{y}(t-\tau)$ in $[t_n;t_{n+1}]$.
    \item Compute the solution $\mathbf{y}(t_{n+1})$ using the interpolant $\Pi_n$ and by stepping the solver
    \item Update the interpolant $\Pi_n$ using the computed solution
    \item End if the interpolant has converged
\end{itemize}
 
The initial guess is usually the extrapolation of the interpolant of the previous step and the end criterion of convergence can vary across cases.

\subsection{Pseudo code for DDE solver}

Following the detailed explanation of the challenges posed by DDE, we present the pseudo code of the DDE solver implemented by  \citet{numerical_dde} that is detailed in Algorithm \ref{al:agl1}, where the general outline of one integration step of a DDE is shown; the DDE solver is illustrated in Algorithm \ref{al:agl2}. For sake of simplicity, we suppose for the pseudo code a single time delay DDE since the general case does not differ from it.

\begin{algorithm}[H]
   \caption{Pseudo code for one DDE numerical integration step}
   \label{alg:one_step}
\begin{algorithmic}[1]
   \STATE {\bfseries Input:} \\ Vector field $\mathbf{f}(t,\mathbf{y}, \mathbf{y}(t-\tau))$ \\ Integration bound $t_n$, $t_{n+1}$ \\ Interpolated estimated solution $\hat{\mathbf{y}}(t)$ in $[t_0;t_{n}]$ \\ Set of detected discontinuities $\Lambda = \{\lambda_{-m}, \dots, \lambda_0 , \dots, \lambda_{r-1} \}$. 
   \IF{$t_{n+1} - t_n > \min(\Lambda) $}
    \STATE{Declare the interpolant $\Pi_n = \hat{\mathbf{y}}$ of $\mathbf{y}(t-\tau)$ in $[t_n;t_{n+1}]$}
   \WHILE{the interpolant $\Pi_n$ has not converged}
    \STATE{Define $\mathbf{f}_{\rm ODE}(t,\mathbf{y}) = \mathbf{f}(t, \mathbf{y}, \Pi_n(t))$}
    \STATE{Step the solver $\mathbf{y}(t_{n+1}) = ODESolve( \mathbf{f}_{\rm ODE}, t_n, t_{n+1}, \hat{\mathbf{y}}(t_n))$}
    \STATE{Update $\Pi_n$ using the computed solution $\mathbf{y}(t_{n+1})$.}
    \ENDWHILE
   \ELSE 
    \STATE{Define $\mathbf{f}_{\rm ODE}(t,\mathbf{y}) = \mathbf{f}(t, \mathbf{y}, \hat{\mathbf{y}}(t-\tau))$}
    \STATE{Step the solver $\mathbf{y}(t_{n+1}) = ODESolve( \mathbf{f}_{\rm ODE}, t_n, t_{n+1}, \hat{\mathbf{y}}(t_n))$}
   \ENDIF 
   \STATE{Determine next time step $t_{\text{next}}$ from solver}
   \IF{step is accepted}
   \STATE{Return updated $\hat{\mathbf{y}}(t)$, next integration bounds $t_{n+1}, t_{\text{next}}$} and $\Lambda$.
   \ELSE 
   \STATE{Check for discontinuities in $[t_n;t_{n+1}]$ i.e} 
   \STATE{$\lambda_r = \min \{ \lambda > \lambda_{r-1}:\lambda$ is a root of odd multiplicity of $g_i(t, y(t)), i \leq r- 1$ \}}
   \STATE{where $ g_i(t, \mathbf{y}(t)) = t - \tau(t, \mathbf{y}(t)) - \lambda_i$}
    \IF{a discontinuity is found, $\lambda_{r+1}$}
    \STATE{Return same $\hat{\mathbf{y}}(t)$, next integration bounds $t_{n}, \lambda_{r+1}$ and $\Lambda \cup \{\lambda_r \}$ .} 
    \ELSE
    \STATE{Return same $\hat{\mathbf{y}}(t)$, next integration bounds $t_{n}$, $t_{next}$ and $\Lambda$.} 
   \ENDIF
   \ENDIF
   \STATE {\bfseries Output:} \\ Interpolated estimated solution \\ Next integration bounds \\ Updated set of discontinuities
\end{algorithmic}
\label{al:agl1}
\end{algorithm}

\begin{algorithm}[H]
  \caption{Pseudo code for DDE solver}
  \label{alg:error_est}
\begin{algorithmic}[1]
  \STATE {\bfseries Input:} \\ Vector field $\mathbf{f}(t,\mathbf{y}, \mathbf{y}(t-\tau))$ \\ Integration bound $t_0$, $t_{F}$ \\ History function $\phi(t)$ \\ Set of history function's discontinuities $\Lambda = \{\lambda_{-m}, \dots, \lambda_0 \}$.
  \STATE{Choose an initial $dt$}
  \STATE{Declare $t_{n}=t_0, t_{n+1}= t_0 + dt$}
  \STATE{Declare interpolated estimated solution $\hat{\mathbf{y}}(t)= \phi(t)$ for $t<t_0$}
  \REPEAT
  \STATE{Algorithm 1}
  \UNTIL{$t_{n+1}=t_F$}
\end{algorithmic}
\label{al:agl2}
\end{algorithm}

\subsection{Memory and computational complexity} 

Neural SDDDEs rely on an ODE solver, thus function evaluations are associated with the same computational cost as NODEs. However, Neural SDDDE has some extra constraints making the method more computationally involved. Hereafter, we compare the complexity of these two schemes; we define $S$ as the number of stages in the Runge-Kutta (RK) scheme used for the time-integration, $N$ the total number of integration steps, and $d$ the state's dimension. 

\paragraph{Memory complexity} DDE integration necessitates keeping a record of all previous states in memory due to the presence of delayed terms. This is because at any given time $t$, the DDE solver must be able to accurately determine $\mathbf{y}(t-\tau)$ through interpolation of the stored state history. This extra amount of extra memory needed depends on the solver used. For example, the additional memory required when using a RK solver for one trajectory is $O(SNd)$. The model's memory footprint is not affected by the number of delays. 

\paragraph{Time complexity} In comparison to NODE, the solution estimate $\widehat{\mathbf{y}}$ needs to be evaluated for each delayed state argument, i.e., $\mathbf{y}(t-\tau_i)$. The cost of evaluating the interpolant is small compared to the cost of computing its coefficients. Similarly, the time complexity is conditioned by the solver used. For example, for a RK scheme, the coefficient computation scales linearly with the number of stages. Hence, Neural SDDDE adds a time cost $O(SD\,d)$ compared to NODE for each vector field function evaluation.

\section{Training information} \label{ap:training_curves}

Table \ref{hyper_ivp} sums up the MLP architecture of each IVP model (i.e., NODE, ANODE and Neural SDDDE) for each dynamical system. ANODE has an arbitrary augmented state of dimension 10 except for the PDE that has 100. Neural Laplace's architecture is the default one taken from the official implementation for all systems. The learning rate and the number of epochs are the same for all models. The optimizer used is \verb?AdaBelief? \citep{zhuang2020adabelief}. Table \ref{number_params} gives the number of parameters for each model. In all of our experiments, we used the Dopri5 solver across all of our models.

\begin{table}[h]
\vskip 0.15in
\begin{center}
\begin{small}
\begin{tabular}{lccccc}
\toprule
 & Width & Depth & Activation & Epochs & $lr$ \\
\hline
Time Dependent DDE & 64 & 3 & relu & 2000 & .001 \\
State Dependent DDE & 64 & 3 & relu & 1000 & .001 \\
Diffusion PDE DDE & 128 & 3 & relu & 500 & .0001 \\
\bottomrule
\end{tabular}
\end{small}
\caption{Model and training hyperparameters}
\label{hyper_ivp}
\end{center}
\end{table}

\begin{table}[H]
\vskip 0.15in
\begin{center}
\begin{small}
\begin{tabular}{lccccr}
\toprule
& NODE & ANODE & Neural DDE & Neural Laplace \\
\hline
Time Dependent DDE & 8513 & 9815 & 8578 & 17194 \\
State Dependent DDE & 8513 & 9815 & 8642 & 17194 \\
Diffusion PDE DDE & 58852 & 84552 & 71653 & 17194 \\
\bottomrule
\end{tabular}
\end{small}
\caption{Number of parameters for each DDE system}
\label{number_params}
\end{center}
\end{table}

\section{Data generation parameters} \label{ap:data_generation}

We expose in Table \ref{hyper_dataset} the parameters used for each dataset generation. The start integration time is always $T_0=0$. $T_F$ refers to the end time integration. \verb?NUM_STEPS? equally spaced points are sampled in $[T_0, T_F]$. The specific delays \verb?DELAYS? and the constant history function $\phi(t)$ function domain are given. Each training dataset is comprised of $256$ datapoints and the testset of $32$ datapoints. We used our own DDE solver to generate the data (Dopri5 solver \citep{dormand} was used.). We then double-checked and compared its validity with Julia’s DDE solver. $\mathcal{U}$ refers to the uniform distribution. For example, Time Dependent DDE's constant history function value is uniformly sampled between 0.1 and 2.0. For the Diffusion Delay PDE the history function value in the column $\phi(t)$ refers to the constant $a$ defined in Section \ref{sec:experiments}.

\begin{table}[H]
\vskip 0.15in
\begin{center}
\begin{small}
\begin{tabular}{lcccccccr}
\toprule
& $T_F$ & num\_steps & delays & $\phi(t)$ \\
\hline
Time Dependent DDE & 20.0 & 200 & $2\sin(t)$ & $\mathcal{U}(0.1, 2.0)$ \\
State Dependent DDE & 10.0 & 150 & $0.5 \cos(\mathbf{y}(t))$ & $\mathcal{U}(0.1, 1.0)$ \\
Diffusion PDE DDE & 4.0 & 100 & $1.0$ & $\mathcal{U}(0.1, 4.0)$ \\
\bottomrule
\end{tabular}
\end{small}
\end{center}
\caption{Dataset generation information}
\label{hyper_dataset}
\end{table}


\newpage
\section{Additional Experimentation}\label{ap:step_history_experiments}
Hereafter, we discuss about the supplementary experiments undertaken to assess the models' quality. In the initial supplementary experiment, we alter the history function $\phi(t)$ which results in modified system behavior, thereby creating new trajectories for section \ref{sec:experiments}'s dynamical systems. This first experiment places each model into a \textit{pure extrapolation regime}; the constant value of the history function $\phi(t)=\mathbf{y}_0$ is sampled outside the range of the training and testing data. This allows to see the models' extrapolation capabilities. The second experiment is more a hybrid approach where the \textit{history function is a step function}:

\begin{equation*}
\begin{aligned}
\phi(t) &= \begin{cases} 
      \mathbf{y}_0 & t\leq t_{\text{jump}} \\
      \mathbf{y}_1 & \text{otherwise}
   \end{cases} \\
t_{\text{jump}} &\sim \mathcal{U}(-\tau_{\text{max}}, 0), \quad \mathbf{y}_0, \mathbf{y}_1 \sim \mathcal{U}(c_0, c_1)
\end{aligned}
\end{equation*}

where $t_{\text{jump}}$ is the largest delay in the system and $c_0, c_1$ are system specific randomly sampled values (see Appendix \ref{sss:history_function} for more details). Not only can the nature of the history step function change but can also have its domain function outside of the training and test data (extrapolation regime).

\begin{figure}[h]
    \centering
    \begin{minipage}{0.45\textwidth}
        \centering
       \includegraphics[width=\textwidth]{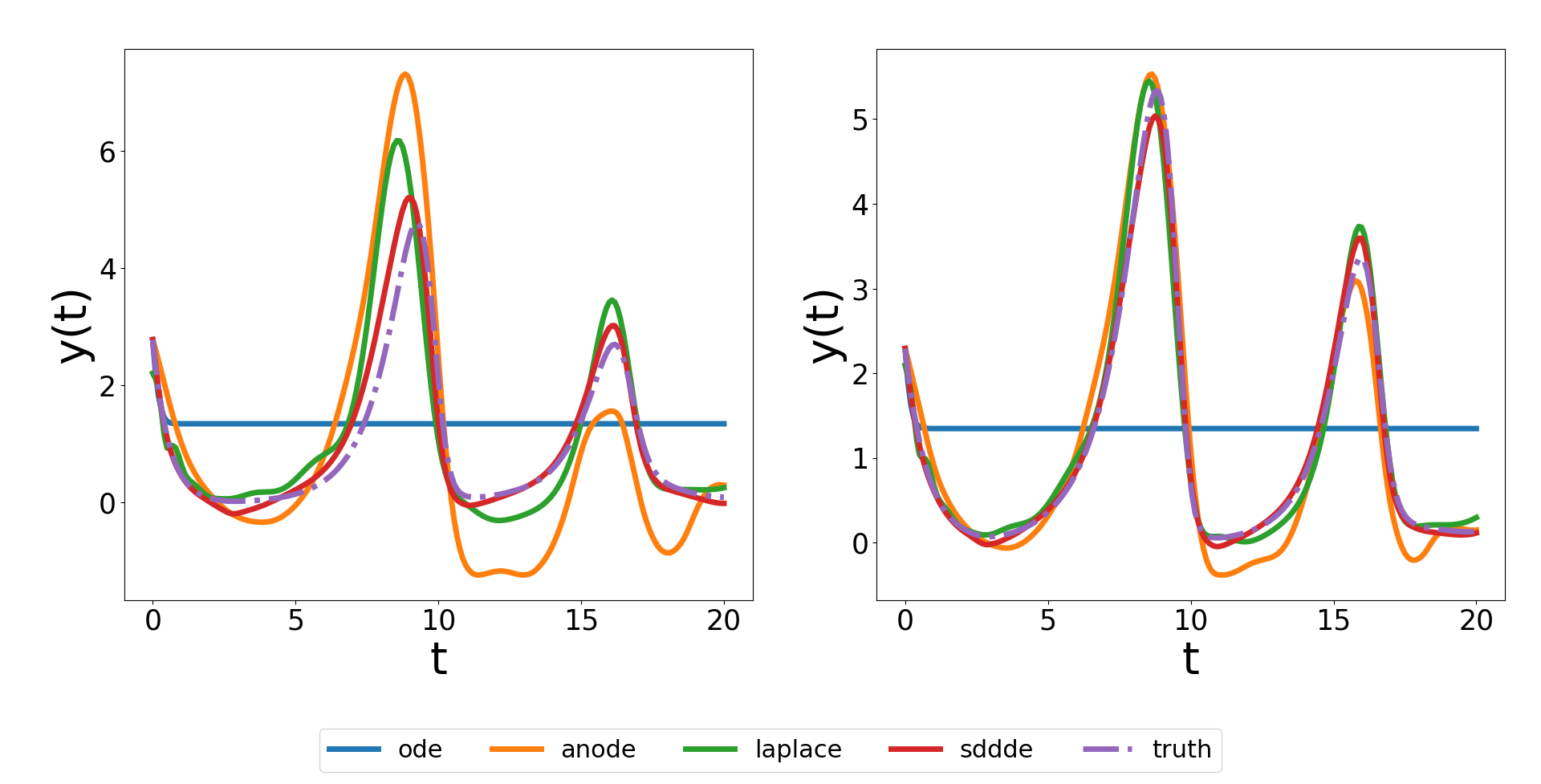}
         \caption{Time Dependent DDE randomly sampled extrapolated trajectory plots }\label{subfig:time_extrapolate}
    \end{minipage}\hfill
    \begin{minipage}{0.45\textwidth}
        \centering
          \includegraphics[width=\textwidth]{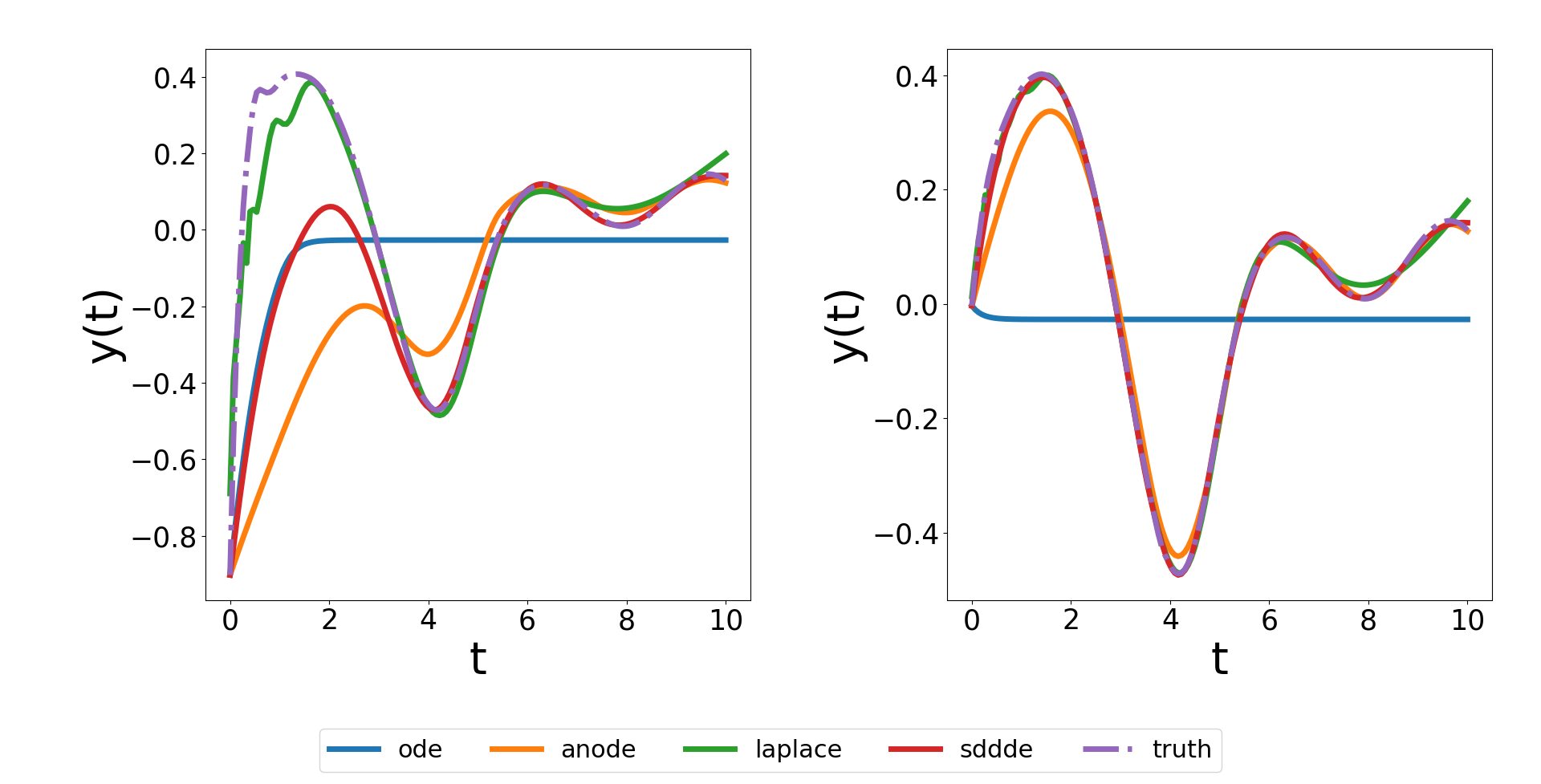}
         \caption{State Dependent DDE randomly sampled extrapolated trajectory plots }\label{subfig:state_extrapolate}
    \end{minipage}
\end{figure}
 
 \begin{figure}[h]
    \centering
     \includegraphics[width=\textwidth, trim={0cm 2.5cm 0 8cm},clip]{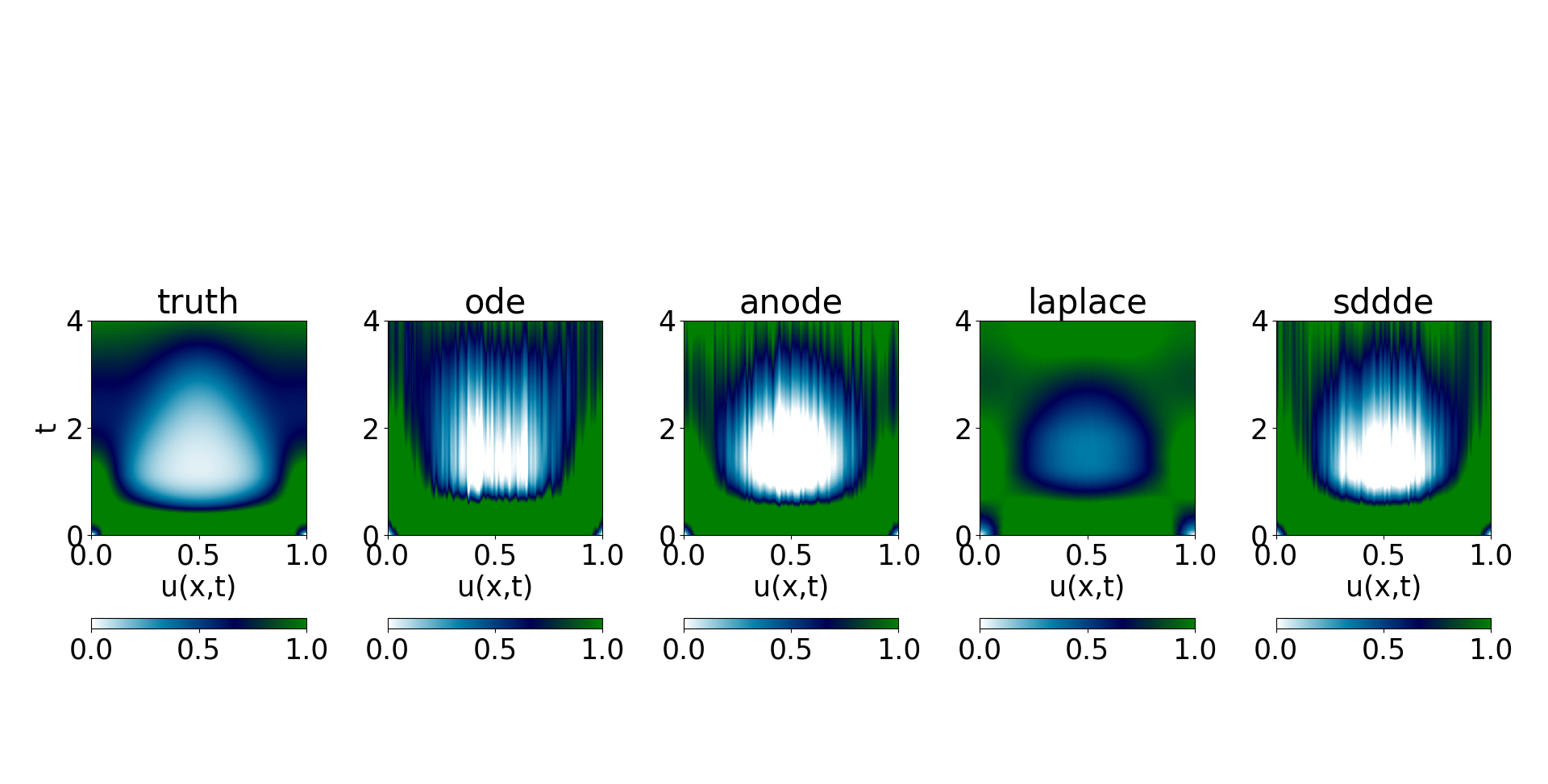}
         \caption{Diffusion Delay PDE  randomly sampled from the extrapolated testset}\label{subfig:diffusion_extrapolate}
    \vspace{1cm}

    \includegraphics[width=\textwidth, trim={0cm 2.5cm 0 8cm},clip]{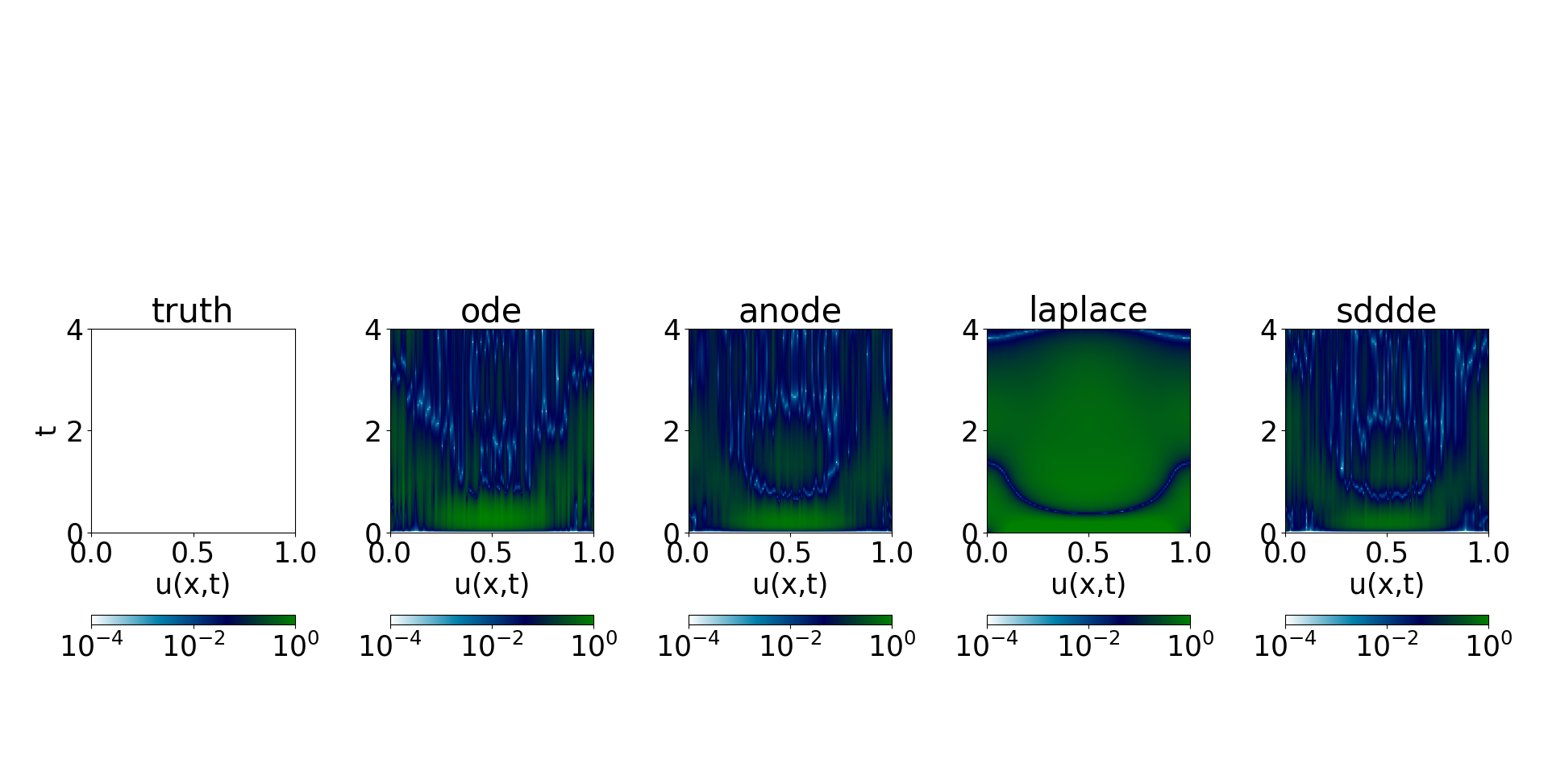}
         \caption{Absolute error of Diffusion Delay PDE randomly sampled from the extrapolated testset}\label{subfig:diffusion_diff_extrapolate}
    
\end{figure}

\paragraph{Extrapolation regime prediction} This experiment really challenges model generalization capabilities. Overall, on certain datasets, some models can extrapolate with new constant history functions that are not too far out from the function domain of history functions used during training; more in details,  trajectories were generated with $\phi(t) = x_0\in [a,b]$: some models are able to exhibit adequate predictions for history functions that have a value near the bounds of $[a,b]$. For the Time Dependent system (Figure \ref{subfig:time_extrapolate}), Neural SDDDE yields better results compared to the other models. NODE produces the trajectory's mean field while ANODE captures the dynamics main trend but with amplitude discrepancies. For the State-Dependent DDE (Figure \ref{subfig:state_extrapolate}), Neural SDDDE once again efficiently captures the dynamics while the other models fail. For the Diffusion Delay PDE displayed in Figure \ref{subfig:diffusion_extrapolate} \& \ref{subfig:diffusion_diff_extrapolate}, overfitting is observed for Neural Laplace. Out of all the IVP models, Neural SDDDE predicts the best possible outcome compared to NODE and ANODE. 

\begin{figure}[H]
    \centering
    \begin{minipage}{0.45\textwidth}
        \centering
         \includegraphics[width=\textwidth]{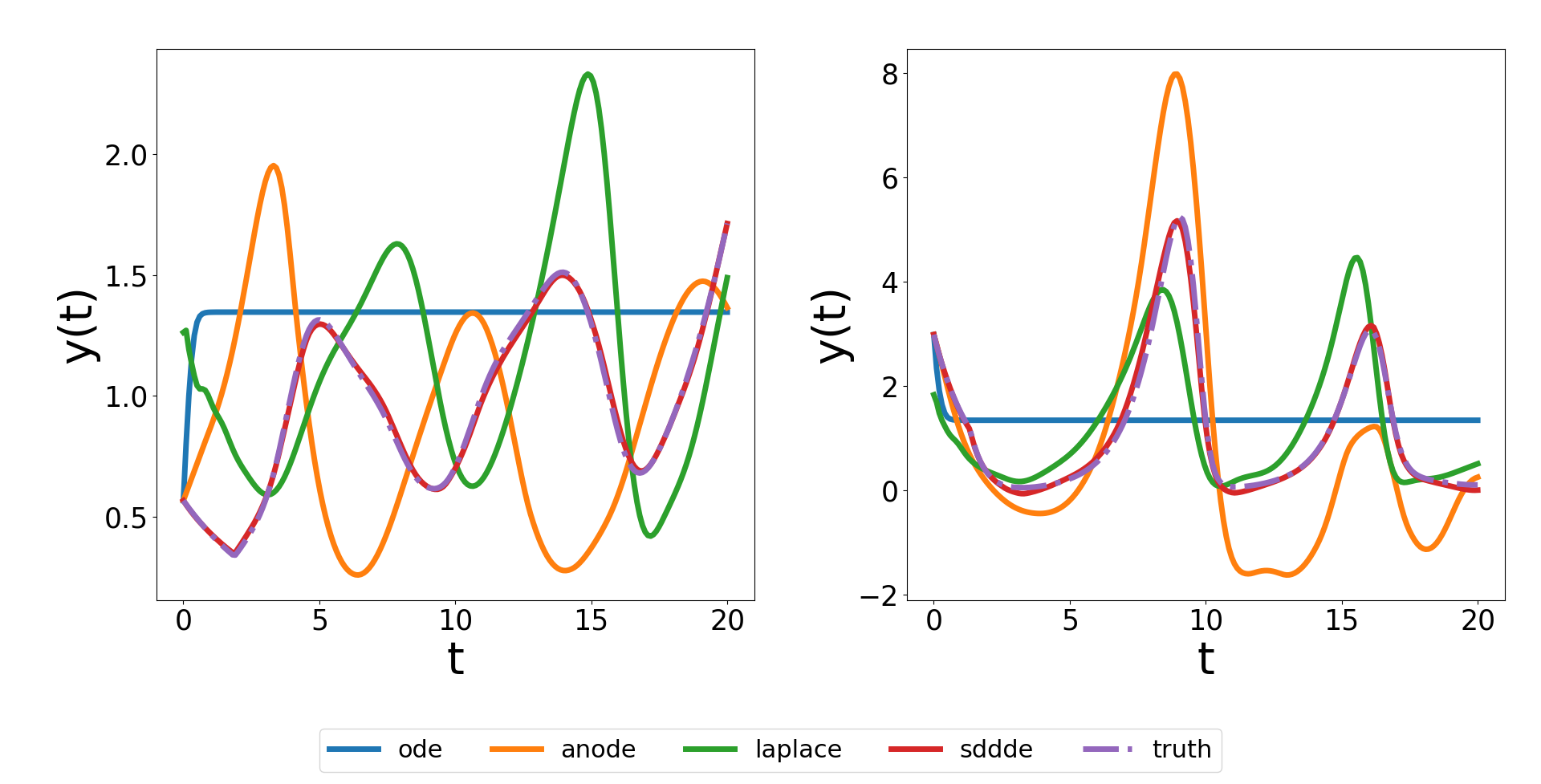}
         \caption{Time Dependent DDE randomly sampled from history step function  }\label{subfig:time_other_history}
    \end{minipage}\hfill
    \begin{minipage}{0.45\textwidth}
        \centering
        \includegraphics[width=\textwidth]{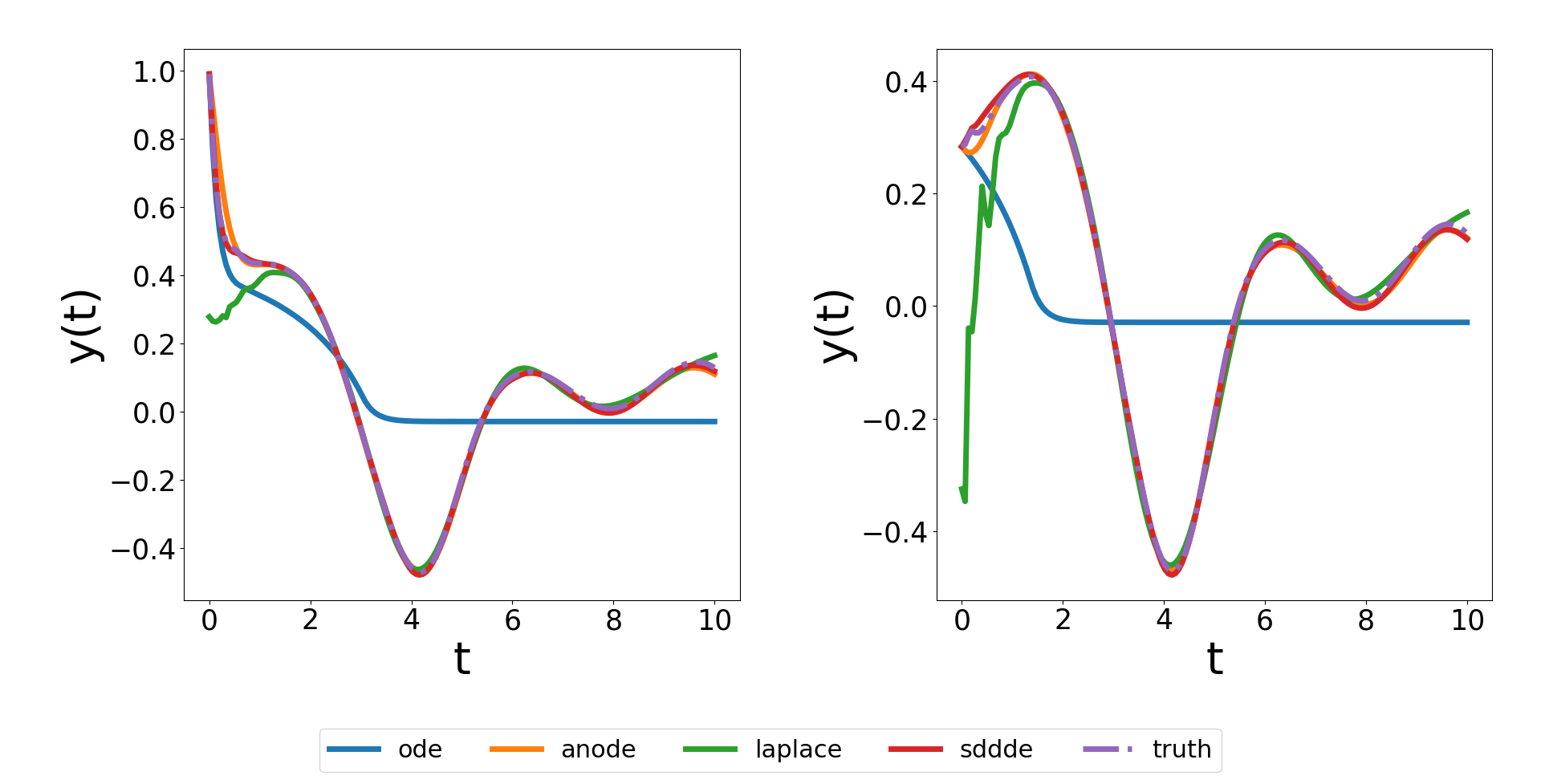}
         \caption{State Dependent DDE randomly sampled from history step function  }\label{subfig:state_other_history}
    \end{minipage}
\end{figure}

\paragraph{Step history function prediction} This third appendix experiment also demonstrates how the modification of the history function leads to changes in the transient regime and impacts later dynamics. Neural Laplace fails to generate adequate trajectories for the Time-dependent DDE system (Figure \ref{subfig:time_other_history}) and the State-Dependent DDE (Figure \ref{subfig:state_other_history}). NODE and ANODE do not generalize well compared to Neural SDDDE that accurately predicts the dynamics. By studying the effect of such a new history function on the Diffusion Delay PDE, we saw that the system's dynamics is not changed substantially, therefore, we decided to omit this system's comparison.

\paragraph{Noise analysis} Finally, we also conduct a noise study on one of the datasets, the Time Dependent DDE system. Each data point is added Gaussian noise that is scaled with a certain factor $\alpha$ of the trajectory's variance. The model is then trained with this noisy data and evaluated on the noiseless testset. In our experiment, we selected 4 scaling factors $\alpha$: $0.02, 0.05, 0.1$ and $0.2$. Results in Table \ref{noiseless_mse} show that our model is robust to noisy data and almost consistently outperforms other models. Additionally, results from Table \ref{noiseless_mse} show that adding a small amount of noise (here $\alpha = 0.02$) makes the learning process more robust, a common result in Machine Learning \citep{noise_robust1,noise_robust2}.

\begin{table}[H]
\begin{center}
\begin{small}
\begin{tabular}{lllll}
\toprule
& NODE & ANODE & Neural Laplace & Neural SDDDE \\
\midrule
$\alpha=0$ & $1.01 \pm .435$ & $.00729 \pm .00235$ & $.0014 \pm .00046$ & $\mathbf{.00148 \pm  .000872}$ \\
$\alpha=0.02$ & $.720 \pm .00254$ & $.0128 \pm .002377$ & $.00881 \pm .00254$ & $\mathbf{.000906 \pm .000441}$ \\
$\alpha=0.05$ & $4.032 \pm 4.225$ & $.03655 \pm .0349$ & $.00977 \pm .00146$ & $\mathbf{.00250 \pm .000951}$ \\
$\alpha=0.1$ & $1.597 \pm 1.100$ & $.0223 \pm .00634$ & $.0154 \pm .00501$ & $\mathbf{.0121 \pm .00534}$ \\
$\alpha=0.2$ & $1.02 \pm .282$ & $.0321 \pm .00319$ & $.0273 \pm .00704$ & $\mathbf{.0186 \pm .00524}$ \\
\bottomrule
\end{tabular}
\end{small}
\caption{Test MSE with the noiseless data averaged over 5 runs of each Time Dependent DDE noise experiments with their standard deviation. Best result bolded.}
\label{noiseless_mse}
\end{center}
\end{table}



\section{Additional Experiment hyperparameters} 
\label{ap:step_history_function}
\label{sss:history_function}

In table \ref{tab:exp_his}  we give the parameters used for each experiment. Extrapolated $\phi(t)$ indicates the possible value of the constant history function (first appendix experiment).  $\tau_{\text{max}}$ and $c_0, c_1$ are described in appendix \ref{ap:step_history_experiments} (second appendix experiment). For the Diffusion Delay PDE, as stated in appendix \ref{ap:step_history_experiments}, the other history step function is omitted. 


\begin{table}[H]
\vskip 0.15in
\begin{center}
\begin{small}
\begin{tabular}{lrccr}
\toprule
& Extrapolated $\phi(t)$ & $\tau_{\text{max}}$ & $c_0$ & $c_1$ \\
\hline
Time Dependent DDE & $\mathcal{U}(2.0, 3.0)$ & $3.0$ & $0.1$ & $3.0$ \\
State Dependent DDE & $\mathcal{U}(-1.0, 0.1)$ & $1/2$ & $-1.0$ & $1.0$ \\
\bottomrule
\end{tabular}
\end{small}
\end{center}
\caption{System specific values for each testing experiment}
\label{tab:exp_his}
\end{table}

\end{document}